\documentclass[sigconf, authorversion, nonacm]{acmart}

\usepackage{multirow}
\usepackage{enumitem, algorithm2e}
\usepackage[utf8]{inputenc} 
\usepackage[T1]{fontenc}    
\usepackage{amsmath, amsfonts}
\usepackage{amsthm}
\usepackage{tcolorbox}
\usepackage{hyperref}
\usepackage{amssymb}
\usepackage{url}
\usepackage{natbib}
\usepackage{bbold}
\usepackage{booktabs}
\usepackage[scale=2]{ccicons}
\usepackage{tikz}
\usetikzlibrary{fadings,shapes.arrows,shadows}  
\usepackage{pgfplots}
\usepgfplotslibrary{dateplot}
\usepackage{xspace}

\makeatletter
\let\oldnl\nl
\newcommand{\nonl}{\renewcommand{\nl}{\let\nl\oldnl}}
\makeatother

\newcommand{\bgu}{\text{ mg/dL}}
\newcommand{\CHO}{\texttt{CHO}}
\newcommand{\BG}{\texttt{BG}}
\newcommand{\INS}{\texttt{INS}}
\newcommand{\idmeal}{\texttt{ID\_meal}}
\newcommand{\CIR}{\texttt{CIR}}
\newcommand{\CF}{\texttt{CF}}

\newcommand{\define}{\overset{.}{=}}

\newcommand{\E}{\mathbb{E}}

\AtBeginDocument{%
  \providecommand\BibTeX{{%
    \normalfont B\kern-0.5em{\scshape i\kern-0.25em b}\kern-0.8em\TeX}}}

\setcopyright{rightsretained}

\begin{document}

\title{Challenging common bolus advisor for self-monitoring type-I diabetes patients using Reinforcement Learning}

\author{Fr\'ed\'eric Log\'e}
\email{frederic.logemunerel@gmail.com}
\affiliation{%
  \institution{Air Liquide R\&D, Jouy-en-Josas, FR}
  \institution{CMAP, Polytechnique, Institut Polytechnique de Paris, Palaiseau, FR}
}

\author{Erwan Le Pennec}
\affiliation{
\institution{CMAP, Polytechnique, Institut Polytechnique de Paris, Palaiseau, FR}
\institution{XPop, Inria Saclay, FR}
}

\author{Habiboulaye Amadou-Boubacar}
\affiliation{\institution{Air Liquide R\&D, Jouy-en-Josas, FR}}


\renewcommand{\shortauthors}{Logé, Le Pennec and Amadou-Boubacar}

\begin{abstract}
Patients with diabetes who are self-monitoring have to decide right before each meal how much insulin they should take. A standard bolus advisor exists, but has never actually been proven to be optimal in any sense. We challenged this rule applying Reinforcement Learning techniques on data simulated with T1DM, an FDA-approved simulator developped by \citep{kovatchev2009silico} modeling the gluco-insulin interaction. Results show that the optimal bolus rule is fairly different from the standard bolus advisor, and if followed can actually avoid hypoglycemia episodes. 

\end{abstract}

\maketitle


\section{Introduction}

\paragraph{Diabetes \& bolus advisor}

Diabetes is a major disease which requires, amongst many things, patients to keep their blood glucose (BG) levels in check. To get an idea of commonly accepted levels: if the BG is lower than 70\bgu, the patient is in hypoglycemia, if higher than 180\bgu, the patient is in hyperglycemia and otherwise in normoglycemia, the ideal is to be around 112.5\bgu. When BG is too low, the patient must take carbohydrates (CHO) to compensate. When it is too high, the patient must take external insulin in order to compensate low pancreatic activity. Insulin intake can be done via an insulin pump which injects continuously varying amounts throughout the day. Most patients however are still injecting punctually, most of the time prior to their meals in order to prepare for the future CHO intake and avoid hyperglycemia.

A standard rule to follow for pre-meal self-injections of insulin is the following piecewise linear formula : for a given measure of blood glucose \BG~ and an intake of carbohydrates \CHO, one should take the following bolus\footnote{Bolus insulin corresponds to rapid insulin injections whereas basal insulin corresponds to long-term insulin.} quantity :
\begin{equation} \label{eq:bolus_advisor}
 \mathbf{bolus}(\CHO, \BG) = \dfrac{\CHO}{\CIR} + \dfrac{\max\{\BG - \BG_{target}, 0\}}{\CF}
\end{equation}
in order to reach the BG target level $\BG_{target}$. In this rule, commonly known as bolus advisor, coefficients \CIR~and \CF~ correspond to the carbohydrate-to-insulin ratio and the correction factor respectively, both being specific to each individual and can be evaluated by medical tests and usually are fluctuant during the day. The first ratio in the rule compensates for CHO intake and the second ratio compensates for excessively high BG levels.

\paragraph{Related works} 
A lot of the research around blood glucose management for diabetes focuses on the artificial pancreas, so the case where the patient is equipped with an insulin pump. The interested reader can find an extensive review here \cite{bothe2013use}. For self-monitoring, \cite{ngo2018control} worked on the best delivery of insulin drugs to facilitate BG management. Based on a complex diabetes simulator, the authors of \cite{daskalaki2013actor} and \cite{sun2018dual} worked on learning adaptively coefficients (\CIR, \CF) which is crucial for deployment.

\paragraph{Objective} 
Although the bolus advisor makes sense, there has not been any study of its performance in its use for patients who are self-monitoring. In this study, we rely on simulated data for type-I diabetes patients and a model-free approach to actually learn the optimal bolus function, tailoring it to meal plans. Using the vocabulary of Reinforcement Learning, we show how to learn non parametric \emph{policies}, the function giving insulin doses based on BG readings and auxiliary information, that outperforms the standard advisor for a given meal pattern.

\section{Methodology}

The problem will be framed as a Markov Decision Process, as presented in the first subsection. The resolution technique, a model-free Reinforcement Learning approach, will be presented next. Finally, we will talk about the use of the simulator available to generate the necessary data for our experiments.

\subsection{Markov Decision Process}

To model the insulin intake process we consider a Markov Decision Process $\mathcal{M} = (\mathcal{S}, \mathcal{A}, \mathbf{R}, \mathbf{T})$ where $(\mathcal{S}, \mathcal{A})$ denote the state and action spaces, $\mathbf{R}$ is the reward function and $\mathbf{T}$ is the transition function. 

\begin{figure}
    \centering
    \includegraphics[width=\linewidth]{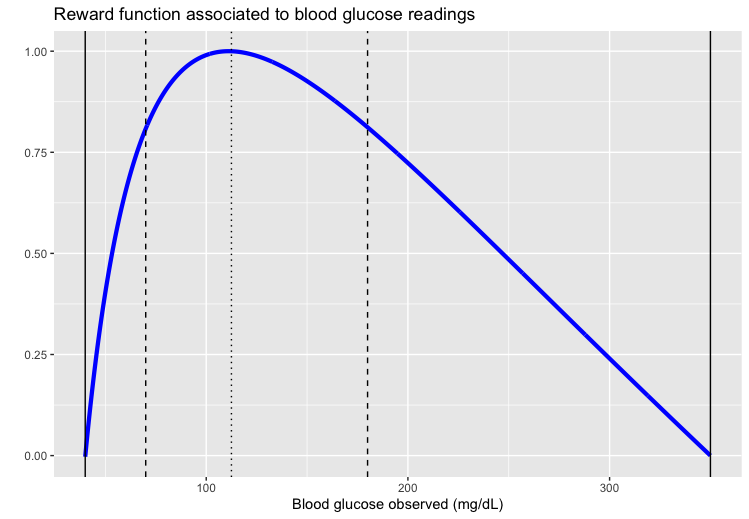}
    \caption{Reward function for BG levels inspired by \cite{kovatchev2000risk}, where the authors proposed to build a function centered at BG level 112.5\bgu~ and symmetric through reference levels, shown with vertical lines.}
    \label{fig:my_reward_function}
\end{figure}

Time steps are set on meals, which follow a strict regular plan. The state variable $S_t$ contains the meal identifier $\idmeal_t$, a categorical variable, and the BG reading $\BG_t$ registered before the meal. The action variable $A_t$ contains the insulin dose $\INS_t$ taken before the meal. Based on the meal plan, we also know the amount of CHO associated with the meal $\CHO_t$, which should simply be a function of $\idmeal_t$. The reward $R_t$ received is a function of $\BG_t$, as represented in figure \ref{fig:my_reward_function} which is indicative of the wellness of the BG profile: the closer it is to 112.5\bgu
~the better.

Let us call \textit{policy} any function $\pi$ which maps $\mathcal{S}$ to $\mathcal{A}$ i.e. any function which to a couple $(\idmeal, \BG)$ assigns an insulin recommendation $\INS$. Writing $\Pi$ the set of policies, we ought to find $\pi^*$, the policy maximizing cumulative rewards:
\begin{equation}
\pi^* = \arg\,\max_{\pi \in \Pi}\, \E_{\mathcal{M}, \pi} \left[ \textstyle \sum_{t = 0}^{+\infty} \gamma^t R_t \right]
\end{equation}
where discount factor $\gamma \in (0,1)$ is chosen by the user and indicates how much long-term rewards matter. Note that the standard bolus advisor from equation \ref{eq:bolus_advisor} belongs to $\Pi$.

\subsection{Q-Learning with function approximation}

For any policy $\pi \in \Pi$ and any couple $(s,a) \in \mathcal{S} \times \mathcal{A}$, consider the state-action value functions
\begin{equation}
Q_{\pi}(s,a) \define \E_{\mathcal{M}, \pi} \left[ \textstyle \sum_{t = 0}^{+\infty} \gamma^t R_t  | S_0 = s, A_0 = a \right].
\end{equation}

Because our state and action spaces are continuous, a common approach to solve the optimization problem is to parameterize the state-action value functions and find the optimal parameters instead. In our case, we will consider the following linear function approximation of $Q(s,a)$ for $s = ({\color{blue} \idmeal},{\color{blue} \BG})$, $a = {\color{blue} \INS}$:
\begin{equation}
Q_{\bar{\alpha}}\left(s, a \right) \define \textstyle \sum_{b = 1}^{B} \sum_{b' = 1}^{B} \alpha_{(b, b')}^{({\color{blue} \text{\idmeal}})} \phi_b^{\BG}({\color{blue} \BG}) \phi_{b'}^{\INS}({\color{blue}\INS})
\end{equation}
with vector parameter $\bar{\alpha}$. The set of functions $\overline{\phi^{\BG}}$ and $\overline{\phi^{\INS}}$ are taken as radial basis functions, centered uniformly on their respective grids. The scaling was made in order to have a certain percentage $p$ of overlap amongst neighbour functions. The number of functions $B$ handles the granularity of the grid.

Linear approximation scheme is widely used in the RL literature, some details can be found in chapter 9 of \cite{sutton2018reinforcement}.

Note that in our context we face the \textit{deadly triad} issue \citep{sutton2018reinforcement} since we jointly apply bootstrap, function approximation and off-policy learning. In order to alleviate this issue, we use a frozen set $\bar{\alpha}^{(frozen)}$ to stabilize target value and apply replay memory to break temporal dependencies between transitions. This strategy has shown good results for instance in \cite{mnih2013playing} where Deep Neural Networks were used as function approximators for Atari game controllers.

\subsection{Gluco-Insulin interactions Simulator}

T1DM simulator (2008 version, accessed throughout Python implementation \citep{simGlucose}) is described in detail in \cite{kovatchev2009silico}. It models the glucose and insulin dynamics taking into account 
\begin{itemize}
\item for glucose : endogenous production, rate of appearance, utilization and renal extraction
\item for insulin : rate of appearance from the subcutaneous tissue and insulin degradation.
\end{itemize}
The simulator takes into account physiology, a given meal scenario and any insulin basal/bolus policy. It also provides good base values for Carbohydrate-to-Insulin Ratio, Correction Factor and Basal Rate for a given physiology.

Although it is built for continuous glucose monitoring, we can use it to generate data in the self-monitoring setting. Using the simulator, the data generated is a set alike
\begin{equation} \label{eq:raw_data_generated}
( \BG_{t}, \CHO_{t}, \INS_{t} )_{t = 1}^{T}
\end{equation}
with a time step of 3 minutes and where the CHO intake depends on the user-specified meal plan and the insulin level corresponds to the bolus insulin which is specified by the user. In our case, the insulin component was sampled uniformly on a proper range of values. The data is then summarized to pre-meal observations as
\begin{equation} \label{eq:transformed_data}
(  \idmeal_t, \BG_t, \CHO_t, \INS_t, R_t )_{t \in \texttt{meal\_times}}.
\end{equation}

\section{Numerical Experiments}

\paragraph{Experiment outline} We proceeded as follows : 
\begin{itemize}
\item Step 0 : choose meal scenario of interest and virtual patient
\item Step 1 : approximate optimal coefficients for the baseline bolus advisor via grid search (policy $\pi_0$)
\item Step 2 : simulate a large amount of data following this meal scenario and explore randomly the insulin intakes within acceptable range, set using the bolus advisor with lower and upper bounds on coefficients (\CIR, \CF)
\item Step 3 : approximate optimal bolus from simulated data with RL approach (policy $\pi_{\overline{\alpha}^*}$)
\item Step 4 : simulate validation data following policies $\pi_0$ and $\pi_{\overline{\alpha}^*}$, compare policy functions and daily glycemic profiles obtained.
\end{itemize}

\paragraph{Meal scenario} We chose the following meal scenario: at 6am the person takes 50 grams of CHO for breakfast. For lunch, at 12am, the person takes 60 grams of CHO, followed by a snack of 15 grams at 3pm. At 8pm, for dinner, the person takes 80 grams of CHO.



\begin{figure}[htbp]
    \centering
    \includegraphics[width=\linewidth]{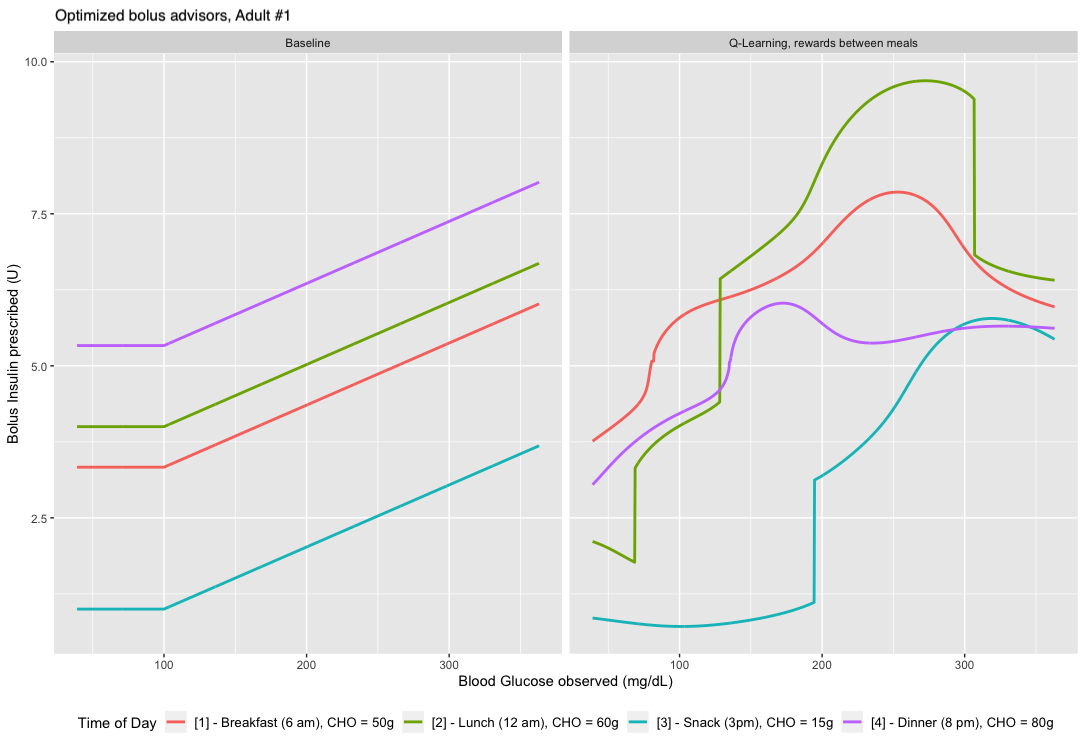}
    \includegraphics[width=\linewidth]{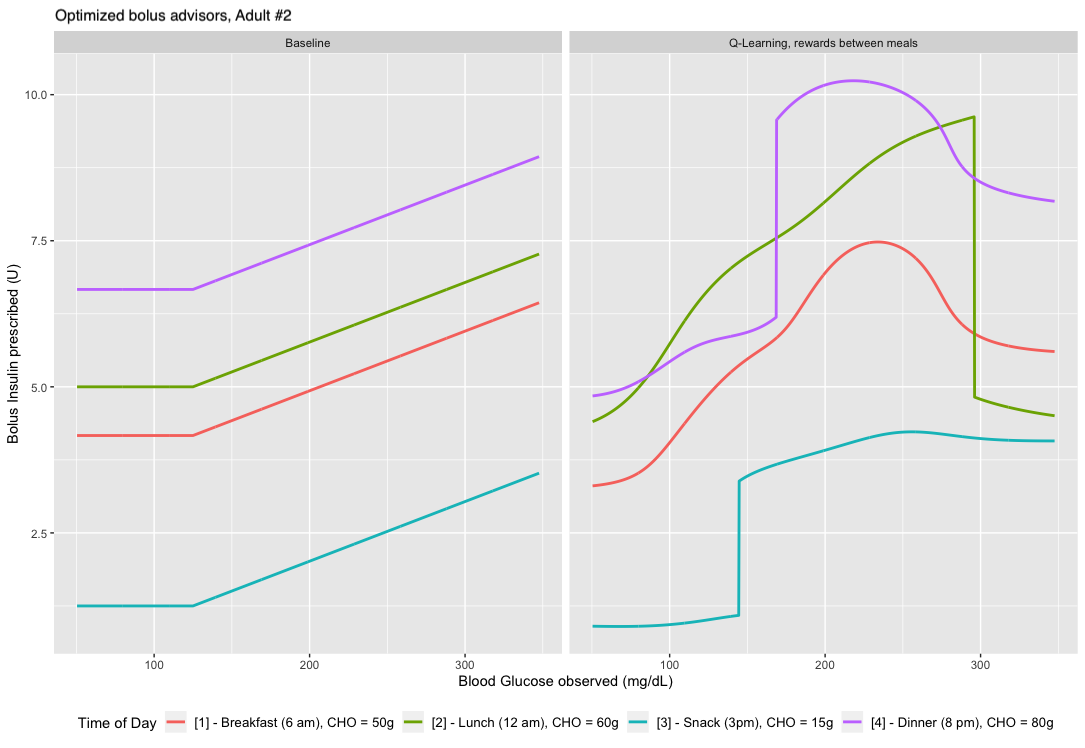}
    \includegraphics[width=\linewidth]{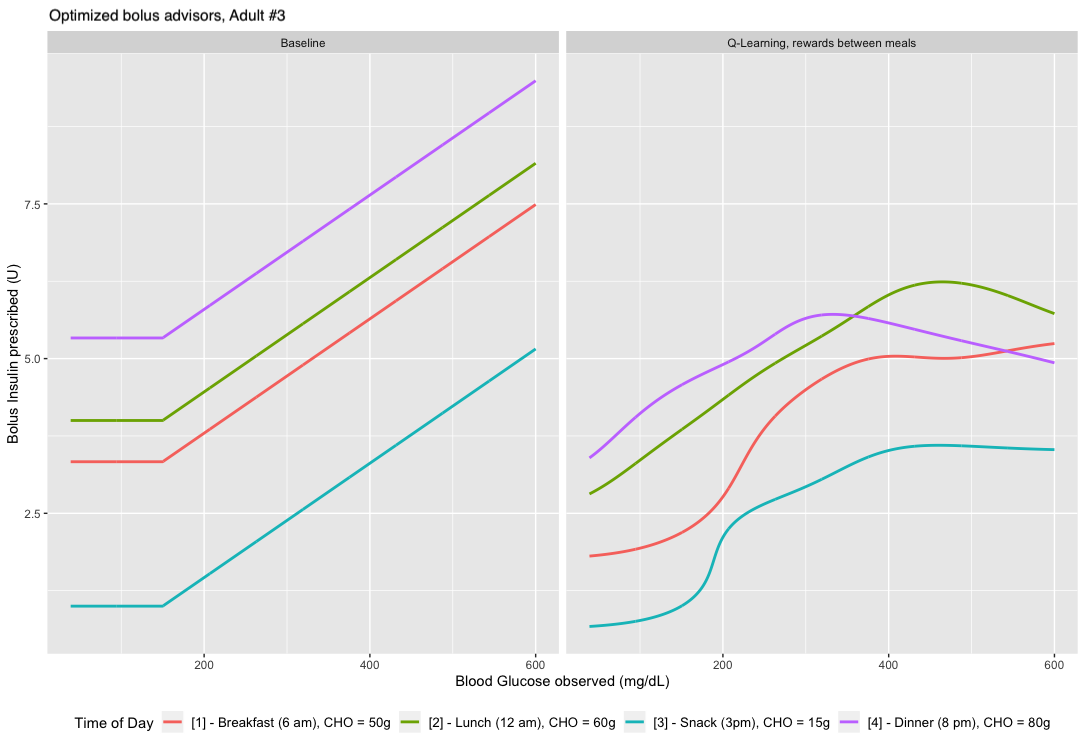}
    \caption{Bolus advisor $\pi_0$ (left) and $\pi_{\bar{\alpha}^*}$ found by Q-Learning with function approximation (right) for each virtual patient (sorted by increasing number). One can observe essentially two things from those graphs : first, it looks as if patient \#1 and \#2 are closer to each other than to patient \#3 in terms of physiology otherwise we would not find the result we did. However the hope to get out a new generic rule seems difficult in this situation. Second, and this is what shows the algorithm took into account sequentiality of the meals and injections : the meal with most carbs does not necessarily require most insulin, which is a fundamental difference from the baseline advisor.}
    \label{fig:bolus_advisors}
\end{figure}

\begin{figure}[htbp]
    \centering
    \includegraphics[height=6cm]{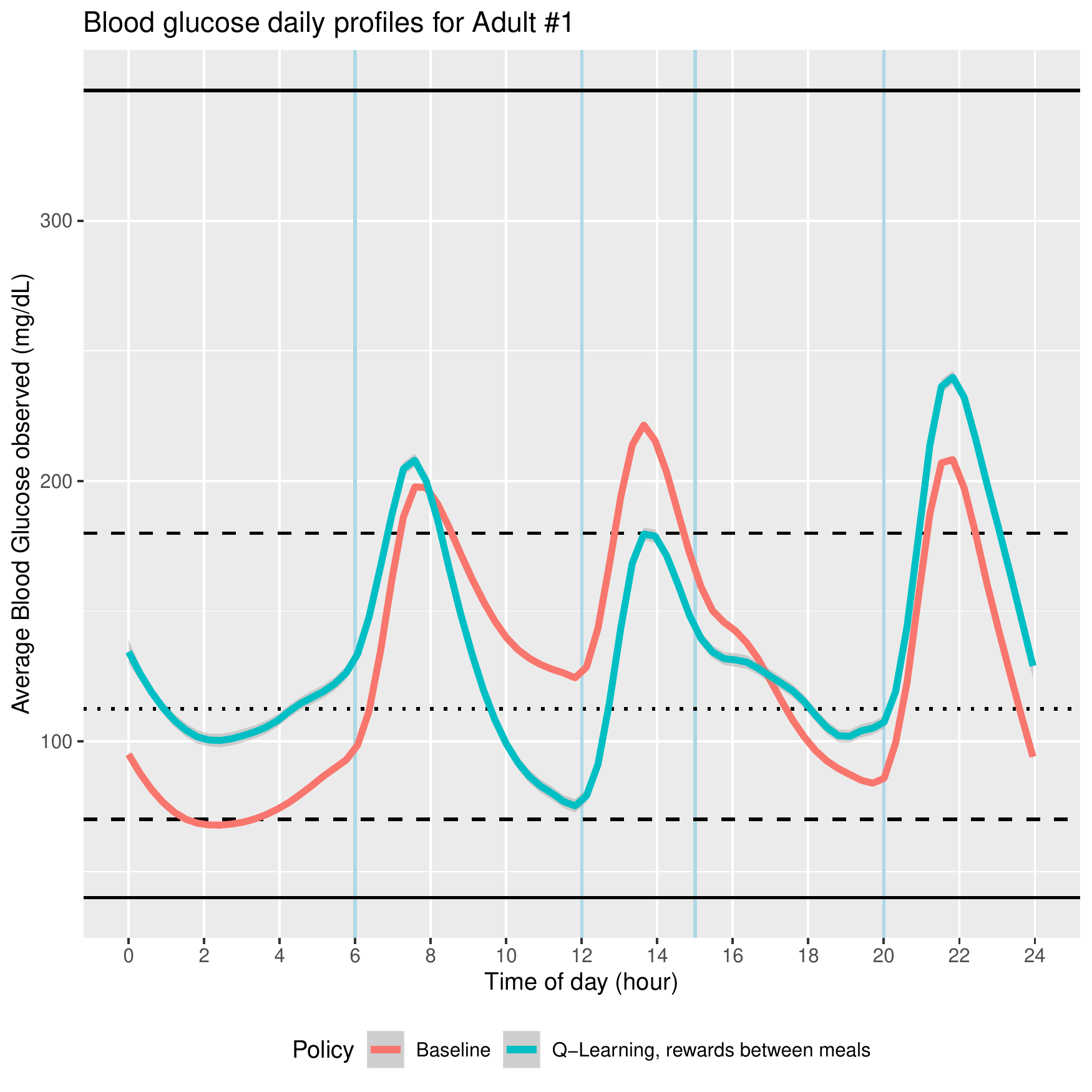}
    \includegraphics[height=6cm]{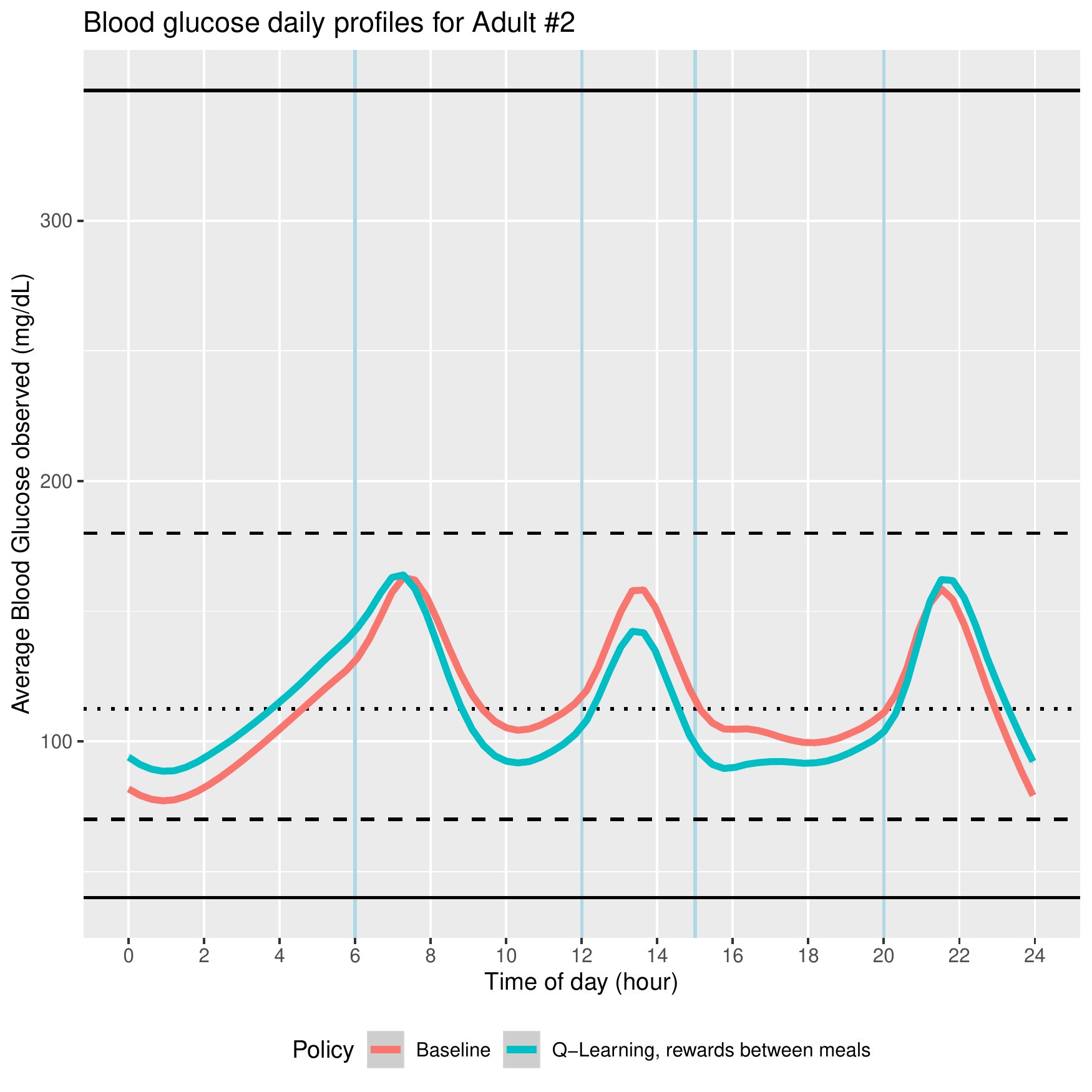}
    \includegraphics[height=6cm]{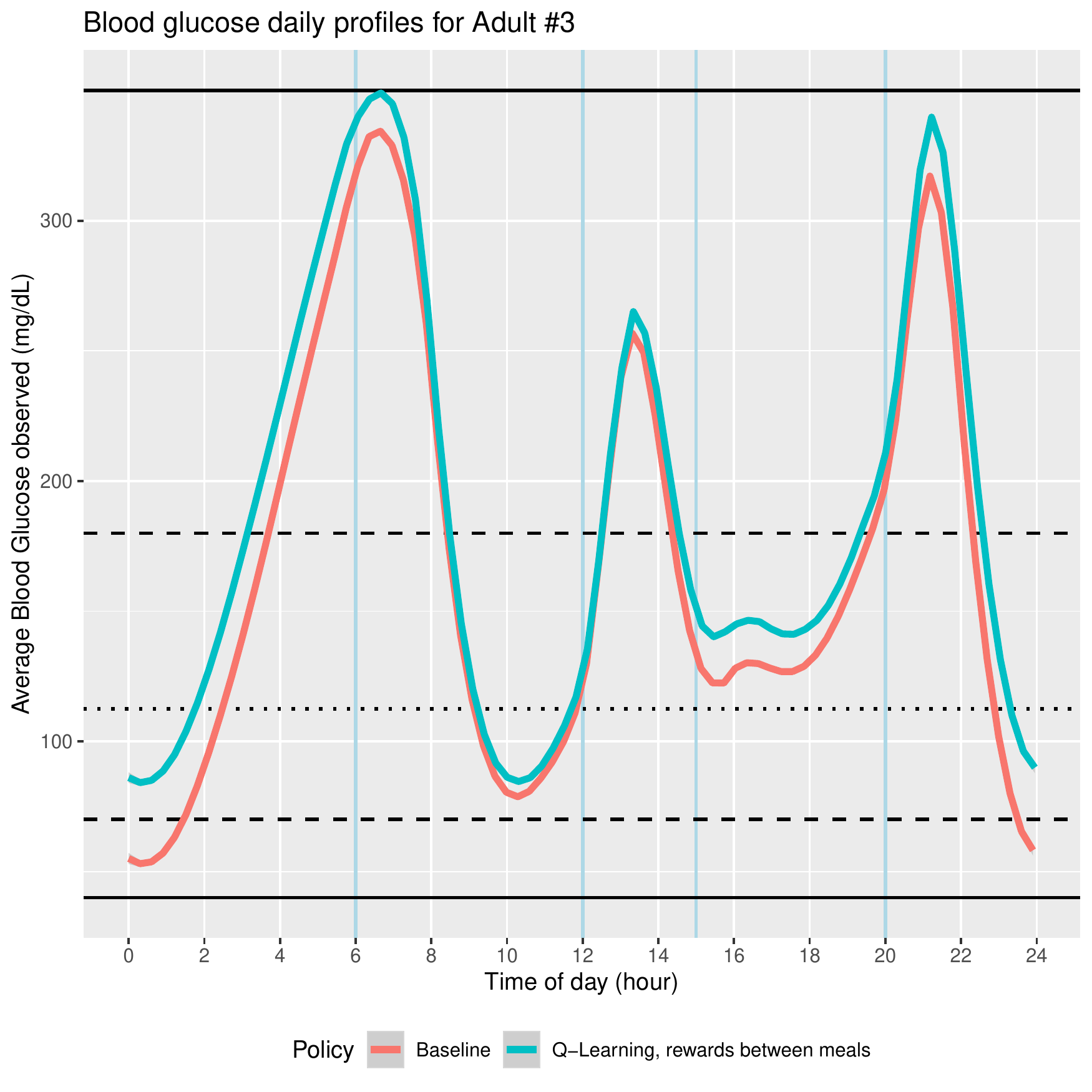}
    \caption{Average BG profiles for the three virtual patients (sorted increasingly by number) for each policy applied, $\pi_0$ in red and $\pi_{\bar{\alpha}^*}$ in light blue. For all adults, the algorithm basically learned to reduce insulin intake at last meal in order to reduce nightly hypoglycemia at the cost of often minor hyperglycemia. The changes are minor in the case of Adult \#2, but in Adult \#3 and \#1 it centers the BG levels around target 112.5\bgu.}
    \label{fig:bg_profiles}
\end{figure}

\paragraph{Simulation details} The procedure above was applied to adults \#1, \#2 and \#3 from simulator T1DM. We simulated for each of them around 8 years of data which amounts to 11680 observations (meals) each. We based our grid search for $\pi_0$ from the following bounds : $\CIR \in [3;30]$, $\CF \in [0.4;2.8]$ and $\BG_{target} \in [100\bgu;150\bgu]$. For the Q-Learning algorithm, we set $B = 8$, giving an 8 by 8 grid of the (\BG, \INS) plane and we set $p = 0.2$ which gave, from our qualitative evaluation, satisfying result. 

Based on the initial data generated, the policies were tested for the same virtual patients (separately, each adult has his/her policy) during 45 days, which is sufficient since the BG enters a stationary state because meal scenario was set deterministic.

\paragraph{Result disposition} In table \ref{tab:metrics}, we included the distribution of BG levels on test data, essentially showing that for adults \#1 and \#3, policy $\pi_{\overline{\alpha}}$ lessens the BG variations outside normoglycemic range compared to $\pi_0$. In figure \ref{fig:bolus_advisors} we present jointly the two policies for each patient, in the left column the standard bolus advisor $\pi_0$ and in the right column the RL-optimized policy $\pi_{\overline{\alpha}^*}$. In figure \ref{fig:bg_profiles} we compare the BG average daily profiles obtained between for the two policies ($\pi_{\bar{\alpha}^*}$ is in blue).

\begin{table}[htbp]
    \centering
    \begin{tabular}{|l rr rr rr|}
    \hline
    \multirow{2}{*}{BG distribution} & \multicolumn{2}{c}{Adult \#1} & \multicolumn{2}{c}{Adult \#2} & \multicolumn{2}{c|}{Adult \#3}  \\
     & $\pi_0$ & $\pi_{\overline{\alpha}^*}$ & $\pi_0$ & $\pi_{\overline{\alpha}^*}$  & $\pi_0$ & $\pi_{\overline{\alpha}^*}$ \\ \hline
    $[40\bgu,70\bgu)$   & .07 & .00 & .00 & .00 & .08 & .00\\
    $[70\bgu,112.5\bgu)$ & .35 & .37 & .53 & .58 & .18 & .20 \\
    $[112.5\bgu,180\bgu)$ & .39 & .47 & .47 & .42 & .35 & .35  \\
    $[180\bgu,350\bgu)$ & .19 & .16 & .00 & .00 & .35 & .39 \\
    $[350\bgu,600\bgu]$ & .00 & .00 & .00 & .00 & .04 & .06 \\ \hline
    \end{tabular}
    \caption{BG distribution for both policies and each patient. Lecture key: for adult \#1, following $\pi_0$, 7\% of BG readings fell within interval [40\bgu, 70\bgu).}
    \label{tab:metrics}
\end{table}


\paragraph{Policies found, figure \ref{fig:bolus_advisors}} Policies $\pi_{\overline{\alpha}^*}$ are quite different in form compared to $\pi_0$ and their form varies with patients. Also, for a given BG level, the recommended bolus does not necessarily increase with \CHO~ amount ; we tend to give more insulin at lunch rather than at dinner, which impacts greatly the BG profile. Typically the patient will be less at risk from nocturnal hypoglycemia which is a well-known issue in the endocrinology community. The policy we proposed properly took into consideration the meal scenario in order to look at future time steps and prevent complications. One may also note that the bolus advisors calibrated sometimes recommend decreasing amounts of insulin with increasing BG level: a great example of that is the lunch recommendations for adults \#1 and \#2 : when BG goes over 300\bgu~ the recommended insulin level drops. This is due to the scarcity of the data collected within this area so it should be considered as a numerical artefact. 

\paragraph{Daily BG profile comparison, figure \ref{fig:bg_profiles}} Let us focus on Adult \#1, whose data is represented in the upper graph. The baseline retained induces low BG during the night (ranging in 70\bgu~ to 100\bgu) whilst it stays quite higher than 112.5\bgu~ from 6am to 3pm. The policy we found (in blue) eliminates night hypoglycemia, traded-off for a two hours period in mild hyperglycemia (<250\bgu) during dinner and a more smoothed-out afternoon. For Adult \#2, the two policies seem quite equivalent. For Adult \#3, we trade-off the night hypoglycemic section by hyperglycemia on the three main meal times. 


\section{Discussion}

\paragraph{Contributions} In this work we challenge a commonly used bolus advisor by relying on virtual patients data, using T1DM simulator \cite{kovatchev2009silico} and a model-free approach from Reinforcement Learning. The policies found bring BG levels closer to reference level of 112.5\bgu~ and in two out of the three adults managed to avoid nocturnal hypoglycemia. These first results attest that a proper bolus advisor does need to be tailored to a patient's physiology (as expressed in the standard bolus advisor through personalized coefficients (\CIR, \CF)), but they should also be tailored to meal and activity plans, which is what we propose here. The code developed for those experiments is available at {\color{blue} \url{https://github.com/FredericLoge/T1DM_qlearning}}.

\paragraph{Future work} We are considering four developments of this work. First, we would like to extend the Reinforcement Learning approach using neural networks as function approximators to take into account historical data properly. Second, instead of optimizing only bolus insulin intake, we could jointly optimize meal and potential activity plans so as to not stress the body with high BG variations. Third, we would like to try different reward functions and compare qualitatively obtained policies. Last but not least, we would like to investigate how quickly we could learn the optimal policies we found here, which did require years of simulated data with randomized insulin intakes, which is obviously unfeasible in real-life. 

\section*{Acknowledgments}

We would like to thank our colleague at Air Liquide, Fouad Megherbi, MD, Medical Director, for his input on this topic and medical references he provided us with to enrich the study.

\section*{Conflict of interest disclaimer}

The author(s) declared no potential conflicts of interest with respect to the research, authorship, and/or publication of this article.

\bibliographystyle{ACM-Reference-Format}
\bibliography{biblio}

\end{document}